\documentclass[conference]{IEEEtran}
\IEEEoverridecommandlockouts
\usepackage{cite}
\usepackage{amsmath,amssymb,amsfonts}
\usepackage{algorithmic}
\usepackage{graphicx}
\usepackage{textcomp}
\usepackage{xcolor}
\usepackage{listingsutf8} 
\usepackage[most]{tcolorbox}
\usepackage{hyperref}

\def\BibTeX{{\rm B\kern-.05em{\sc i\kern-.025em b}\kern-.08em
    T\kern-.1667em\lower.7ex\hbox{E}\kern-.125emX}}
\begin{document}
\makeatletter
\newcommand{\newlineauthors}{%
  \end{@IEEEauthorhalign}\hfill\mbox{}\par
  \mbox{}\hfill\begin{@IEEEauthorhalign}
}
\makeatother

\title{Probabilistic Safety Verification for an Autonomous Ground Vehicle: A Situation Coverage Grid Approach\\
\thanks{Acknowledgment: This work was supported by the Centre for Assuring Autonomy, a partnership between Lloyd’s Register Foundation and the University of York (https://www.york.ac.uk/assuring-autonomy/).}
}
\author{\IEEEauthorblockN{ Nawshin Mannan Proma}
\IEEEauthorblockA{\textit{Dept. of Computer Science} \\
\textit{University of York}\\
York, United Kingdom \\
nawshinmannan.proma@york.ac.uk}
\and
\IEEEauthorblockN{ Gricel Vázquez }
\IEEEauthorblockA{\textit{Dept. of Computer Science} \\
\textit{University of York}\\
York, United Kingdom \\
gricel.vazquez@york.ac.uk}
\and
\IEEEauthorblockN{ Sepeedeh Shahbeigi}
\IEEEauthorblockA{\textit{Dept. of Computer Science} \\
\textit{University of York}\\
York, United Kingdom \\
sepeedeh.shahbeigi@york.ac.uk}
\newlineauthors
\IEEEauthorblockN{ Arjun Badyal}
\IEEEauthorblockA{\textit{Dept. of Computer Science} \\
\textit{University of York}\\
York, United Kingdom \\
arjun.badyal@york.ac.uk}
\and
\IEEEauthorblockN{ Victoria Hodge}
\IEEEauthorblockA{\textit{Department of Computer Science} \\
\textit{University of York}\\
York, United Kingdom \\
victoria.hodge@york.ac.uk}
}

\maketitle
\begin{abstract}
As industrial autonomous ground vehicles are increasingly deployed in safety-critical environments, ensuring their safe operation under diverse conditions is paramount. This paper presents a novel approach for their safety verification based on systematic situation extraction, probabilistic modelling and verification. We build upon the concept of a situation coverage grid, which exhaustively enumerates environmental configurations relevant to the vehicle's operation. This grid is augmented with quantitative probabilistic data collected from situation-based system testing, capturing probabilistic transitions between situations. We then generate a probabilistic model that encodes the dynamics of both normal and unsafe system behaviour. Safety properties extracted from hazard analysis and formalised in temporal logic are verified through probabilistic model checking against this model. The results demonstrate that our approach effectively identifies high-risk situations, provides quantitative safety guarantees, and supports compliance with regulatory standards, thereby contributing to the robust deployment of autonomous systems. 
\end{abstract}


\section{Introduction}

Industrial Autonomous Ground Vehicles (AGVs) are expected to provide service across various domains of human life. AGVs are designed to provide services such as delivery, transportation, manufacturing and warehouse storage~\cite{PatkarMehendale2025}. Industrial AGVs are self-navigating systems that use sensors such as LiDAR, radar, cameras, and GPS for environmental perception. They incorporate AI and machine learning for data processing and decision-making, and rely on advanced control systems, to operate accurately and independently in complex environments~\cite{PatkarMehendale2025}. The benefits of using AGVs in industry include the ability to serve a wider community through enhanced automation. However, to realise their full potential, their safety should be ensured. One of the aspects of ensuring AGVs' safety is to consider the range and diversity of their operating contexts, and incorporate this information into the safety requirements and subsequent verification of the AGV system.   


Testing autonomous software differs fundamentally from traditional software testing, as autonomous systems must operate in dynamic, unpredictable environments and respond to a wide variety of real-world situations such as interactions with other vehicles, obstacles, and humans~\cite{Rob2015}. To ensure that all the variations are considered while developing and testing these systems, numerous coverage techniques including system coverage, requirement coverage, and scenario coverage have been explored in autonomous software testing \cite{kurakin2018adversarial,pei2017deepxplore,tian2018deeptest,ulbrich2015defining,abdessalem2018testing,iqbal2015environment}. However, Existing techniques often fail to capture this situational diversity, making it challenging to ensure the safety and reliability of AGVs in practice. Nevertheless, autonomous software for AGVs is hard to test as it can face a wide range of external situations in an unstructured environment (e.g. encountering obstacles, other AGVs and humans on its path). 
To cover dynamic scenarios arising from the interaction of an AGV and its environment, the situation coverage \cite{Rob2015,Hawkins2019} approach is a promising technique adapted for the testing of autonomous AGVs under a broad range of possible situations. The situation coverage approach identifies potential environmental factors the system may encounter, determines how they can change, and ensures that both individual factors and their combinations are tested.   
Existing work on situation coverage(~\cite{Tahir2022,nawshin2023,nawshin2024, proma2025scaloft}) does not consider uncertain behaviour that may occur when a robotic system abruptly changes from one situation to another, nor provide probabilistic guarantees of the safety of situations when such changes exist. To improve safety verification for AGVs, our approach combines a situation coverage grid, extracted from the Operational Domain Model (ODM)\cite{hawkins2022guidance}, with the probabilistic system's behaviour data, and leverages probabilistic model checking to provide safety guarantees based on the requirements derived from the hazard analysis.

The main contributions of this paper are as follows:
\begin{itemize}
    \item A structured methodology for constructing and augmenting a situation coverage grid from an Operational Design Domain (ODM), enabling systematic analysis of environmental variability.
    
    \item A novel integration of situation coverage with probabilistic modelling, capturing both condition-specific risks and transitions to unsafe states within a unified framework.
    
    \item A formal approach to specifying and verifying probabilistic safety requirements using temporal logic, bridging the gap between qualitative hazard analysis and quantitative verification.
    
    \item An automated synthesis of system-level probabilistic models from empirical data and augmented coverage structures, enabling quantitative safety verification.
    
    \item A representative case study demonstrating the practical applicability and effectiveness of the approach in identifying and analysing high-risk situations.
\end{itemize}


\begin{figure}
    \centering
    \includegraphics[width=.8\linewidth]{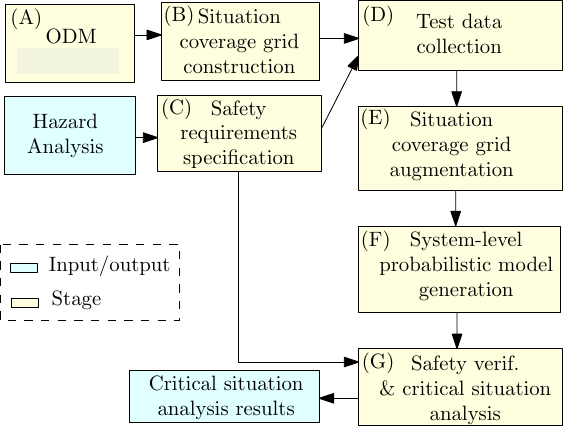}
    \caption{Overview of our approach for the quantitative probabilistic verification of situations under safety requirements, from hazard analysis results and ODM definition.}
    \label{fig:approach}
\end{figure}


\section{Background}
We briefly introduce key concepts underpinning our methodology in this section.

\subsection{Situating Our Methodology in the V-Cycle}
Our method addresses system-level V\&V in the V-cycle. Safety requirements are derived from hazard analysis (e.g., HARA(Hazard And Risk Analysis), HAZOP(HAZard and Operability study), STPA(Systems Theoretic Process Analysis)~\cite{molloy2024hazard}) and the Operational Design Domain (ODM), then formalised using probabilistic temporal logic. These are verified against an empirically grounded Discrete-Time Markov Chain (DTMC) model generated from an augmented situation coverage grid. This supports quantitative safety analysis under diverse operational conditions. A situation is defined by a tuple of binary environmental variables.

\subsection{ODM}
The \textbf{ODM} defines where, when, and under what conditions an autonomous system, such as an AGV, is designed to operate safely and effectively\cite{hawkins2022guidance}. It specifies the set of conditions under which the system can function as intended, and beyond which it may not be able to ensure safety or optimal performance. Fig.~\ref{fig:odm} depicts a portion of an ODM, where objects are categorised as either dynamic or static. These categories are further refined based on specific characteristics, represented as leaf nodes in the hierarchy.
\subsection{Situation coverage grid}
To systematically evaluate the AGV's behavior in diverse operational contexts, we then construct a discrete situation coverage grid.

\textit{Definition 1 (Situation)}. Given a situation hyperspace--with each leaf representing a finite countable set of values--a \textit{situation} $s$ is defined by a list with the combination of its leaves' values. We refer to the set of all situations as $S$. Fig.~\ref{fig:hyperspace} demonstrates such a hyperspace.

\subsection{PCTL}
Probabilistic Computation Tree Logic (PCTL) is a logic language for the description of probabilistic formulas~\cite{ciesinski2004probabilistic}. It allows the specification of unambiguous properties and their further automated analysis using model checker tools, such as PRISM~\cite{kwiatkowska2022probabilistic,kwiatkowska2011prism}. In this stage, we describe quantitative properties such as the probability of an unsafe event. For example, "\textit{what is the probability of an AGV collision with an static obstacle}", written as $\mathcal{P}_{=?}[F\ \text{"collision\_static"}]$, where $P$ is the probabilistic operators, $F$ is the "eventually" logic operator.  

\textit{PCTL definition}~\cite{ciesinski2004probabilistic}. Given the set $AP$ of atomic propositions, the PCTL syntax is defined based on three types of constructs: 
(i) \emph{state formulae:} 
$\phi ::= \textit{true} \mid\ a \mid\; \neg \phi \;\mid\; \phi \land \phi,$ 
where $a \in AP$ is an atomic preposition; 
(ii) \emph{ path formulae:} 
$\psi ::= \mathsf{X}\,\phi \;\mid\; \phi\,\mathsf{U}^{\le k}\,\phi \;\mid\; \phi\,\mathsf{U}\,\phi$, where $k \in \mathbb{N}$ is a time bound; 
and (iii) \emph{qualitative queries:} 
$\Phi ::= P_{\bowtie p} [\psi]$, 
where $\bowtie \in \{\!<,\le,>,\ge\!\}$, $p \in [0,1]$, and \emph{quantitative queries} $P_{=?}[\psi]$. 

Derived operators such as "eventually" (F) can be defined using other temporal operators; e.g., F $\phi$ is defined as true $\mathsf{U}$ $\phi$. An example of a \emph{quantitative} query is $\mathcal{P}_{=?}[F \text{"success"}]$, which computes the probability of eventually reaching a state labelled "success". Filters in PRISM can evaluate a property from a specific initial state $s'$. These take the form: {\sf filter(state}, $\Phi$, $s'${\sf)}.

\section{Methodology}

In this section, we present our approach for the safety verification and analysis of probabilistic models obtained from situation coverage grid. Our approach consisting of stages A-G as depicted in  Fig.~\ref{fig:approach}. 
We continue presenting each stage. 


\textbf{ODM definition (A)}. Inspired by de Gelder et al. \cite{deGelder2024Coverage}, our approach adopts the ISO 34504’s~\cite{iso34504} tag-based scenario categorisation, as it provides a standardised and interoperable taxonomy for decomposing an AGV’s ODM into a high-dimensional space that can be systematically and reproducibly explored. 
We refer to this space as the situation hyperspace \cite{proma2025scaloft}. This conceptual framework serves as an initial study to organize key dimensions such as environmental features and navigation context in a structured manner. Fig.~\ref{fig:hyperspace} illustrates how the navigation context can include elements such as door traversal. Leaves are associated to a finite set of values. For example, door traversal can be True when the AGV passes through a door, and False otherwise.

The level of detail determined by domain experts and safety experts according to the specific requirements of the application domain. The hyperspace is designed to be extensible, allowing additional axes to be incorporated as needed. This enables a structured and comprehensive representation of the possible combinations of operating scenarios that an AGV may encounter within its ODM, thereby facilitating more robust and coverage driven safety analysis.

\begin{figure}[h]
    \centering
    \includegraphics[width=0.4\textwidth]{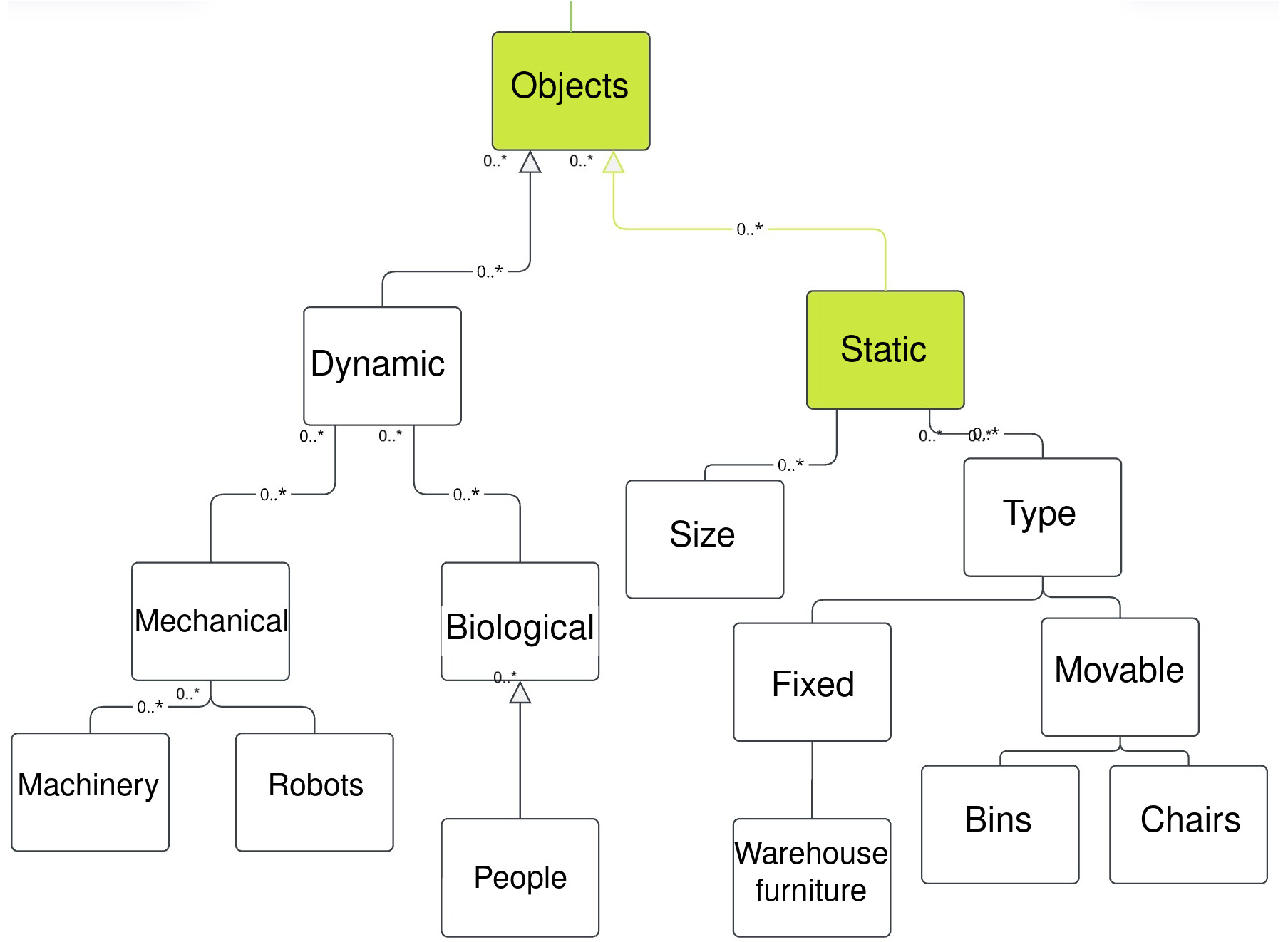}
    \caption{Portion of an ODM showing objects found in an AGV's warehouse setting example. Full ODM can be found in our GitHub repository~\cite{github2025}}.
    \label{fig:odm}
\end{figure}

\begin{figure}[h]
    \centering
    \includegraphics[width=0.4\textwidth]{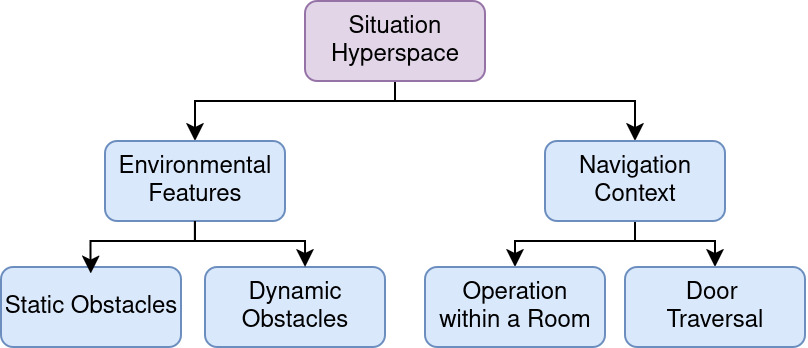}
    \caption{Transformation of the ODM from Fig.~\ref{fig:odm} into its situation hyperspace (purple). The axis layer (blue) defines key factors. Combinations of leaf nodes result in different situations.}
    \label{fig:hyperspace}
\end{figure}

\textbf{Situation coverage grid construction (B)}. In this work we present the situations as tuples: presence of a door (\(d \in \{0,1\}\)), presence of an obstacle on the path (\(o \in \{0,1\}\)), presence of a human (\(h \in \{0,1\}\)), and presence of another AGV (\(a \in \{0,1\}\)) (see Table~\ref{Table:Sit_Grid}). This yields a finite set of situations, \( S = \{s_1, s_2, \dots, s_n\} \). Each element \( s_i \in S \) represents a unique configuration of the environment, enabling the construction of a structured grid for its use in exhaustive testing of the AGV’s navigation and decision-making mechanisms. 

\textbf{Safety requirements specification (C)}. In this stage, system-level safety requirements are extracted and defined using HARA and requirements engineering. Extracting $probabilistic$ requirements for the AGV is a non-trivial task, often hindered by the specialised expertise needed for accurate description and analysis. As a result, such requirements are frequently overlooked or inadequately addressed. When they are specified, they are typically expressed in natural language, which introduces potential ambiguity. To overcome this limitation, in stage C, the extracted requirements are also formalised using PCTL, enabling rigorous and unambiguous specification suitable for formal verification. This formalisation marks a crucial step toward ensuring that the AGV's behavior can be systematically analysed and validated against its safety objectives.

\textbf{Test data collection (D)}. So far we have the coverage grid consisting of the situations which are then considered as system states in the verification stage. Data is collected from (1) normal behaviour of the AGV under different situations; (2) changes between situations; and (3) safety violations. The latter is relevant as we are interested on the analysis of safety properties identified from stage C.

The AGV is deployed (or simulated) in situations aligned with each scenario \(s\), and relevant data are collected, such as trajectory information, decision outputs, interaction logs with dynamic elements (e.g., humans or obstacles), and safety-critical events such as collisions. The empirical probability of a safety violation in each situation $s$, such as the probability of a collision $P_{\text{collision}}(s)$, is estimated using either frequentist analysis or Bayesian approaches~\cite{kwiatkowska2022probabilistic}. This enables a quantitative assessment of system safety and supports the identification of high-risk scenarios.

\textbf{Situation coverage grid augmentation (E)}.  To capture the dynamic nature of real-world environments, the situation coverage grid is augmented with probabilistic transitions between situations. This uncertainty augmentation step is central to our approach. Transitions reflect how one situation may evolve into another due to changes in environmental factors, such as the sudden appearance of a human or another AGV. Transition probabilities are derived from empirical observations from (D) and modelled using probabilistic distribution functions, describing the AGV’s behaviour under uncertain and changing conditions. As we are also interested in the analysis of elicited safety properties, transitions of hazard or failure states are also part of the transitions from each situation (e.g., an AGV colliding with a person is detected when the system is tested from a given situation).



\textbf{System-level probabilistic model generation (F)}. A probabilistic Discrete Time Markov Chain (DTMC) model is automatically generated from the augmented situation coverage grid in stage E. 
The states of the generated DTMC represent each of the situations. An additional critical state is defined for the system's failure, where the safety of the system is theoretically guaranteed as long the system avoid such state. State transitions are automatically fetched from the augmented situation coverage grid. A transition between two states exist if their probability is greater than zero, i.e., it is defined in the grid. The failure state is an absorbing state as it has no outgoing transitions to other states, representing a catastrophic failure of the system.

\textit{Formal definition}. The DTMC model of an augmented situation coverage grid is defined by the tuple $\mathcal{D} = (S, s_i, \delta, AP, L)$ where:
\begin{itemize}
    \item $S'$ is the finite set of states representing situations and failures; 
    \item $s'_i \in S'$ is an initial situation from which the analysis of a safety property is performed;
    \item $\delta : S' \rightarrow \mathit{Dist}(S')$ is the probabilistic transition function defined for each state $s'\in S'$ over $S$; 
    \item $L : S' \rightarrow 2^{AP}$ labels each state with atomic prepositions (e.g. $s_1$ for situation 1 and $Fail$ for the failure state, where $s_1,Fail\in AP$). 
\end{itemize}

\textbf{Safety verification (G)}. The generated probabilistic situation model is formally verified against safety properties identified through prior safety specifications in stage C, using probabilistic model checking (PMC). PMC is a formal verification technique that systematically explores the state space of the probabilistic model to assess the likelihood of a given property defined in PCTL from an initial situation. Tools like PRISM and Storm~\cite{storm} automate this process by leveraging a repository of efficient algorithms. This quantitative verification enables the identification of high-risk situations by computing its probability of failure. Such analysis informs decision-making by highlighting situations that exceed acceptable risk thresholds, thus supporting explainability and system adaptation. An example is provided in Section~\ref{sec:E System verification}. 
This stage ensures that safety requirements are not only traceable but also quantitatively validated, thereby strengthening the overall assurance case for safe deployment. This quantitative verification of probabilistic behavior across situations represents one of the most impactful and distinguishing aspects of our approach, setting it apart from traditional testing and qualitative methods.

\section{Preliminary Assessment: AGV Warehourse Scenario}

In this section, we use an AGV warehouse scenario to illustrate and assess the applicability of our proposed approach for the quantitative safety verification of identified situations. All supplementary materials are available on our GitHub~\cite{github2025}. 

\textbf{Case study description.} The motivating case scenario consist of an AGV patrolling a given set of locations in a warehouse illustrated in Fig.~\ref{fig:case_study}. These locations are in separate rooms, connected through automated doors that open when the AGV is close to them. Other AGVs are performing different tasks around the warehouse, such as pick-and-place activities and delivering packages. The AGV might also encounter human workers within the warehouse. 

\begin{figure}
    \centering
    \includegraphics[width=.8\linewidth]{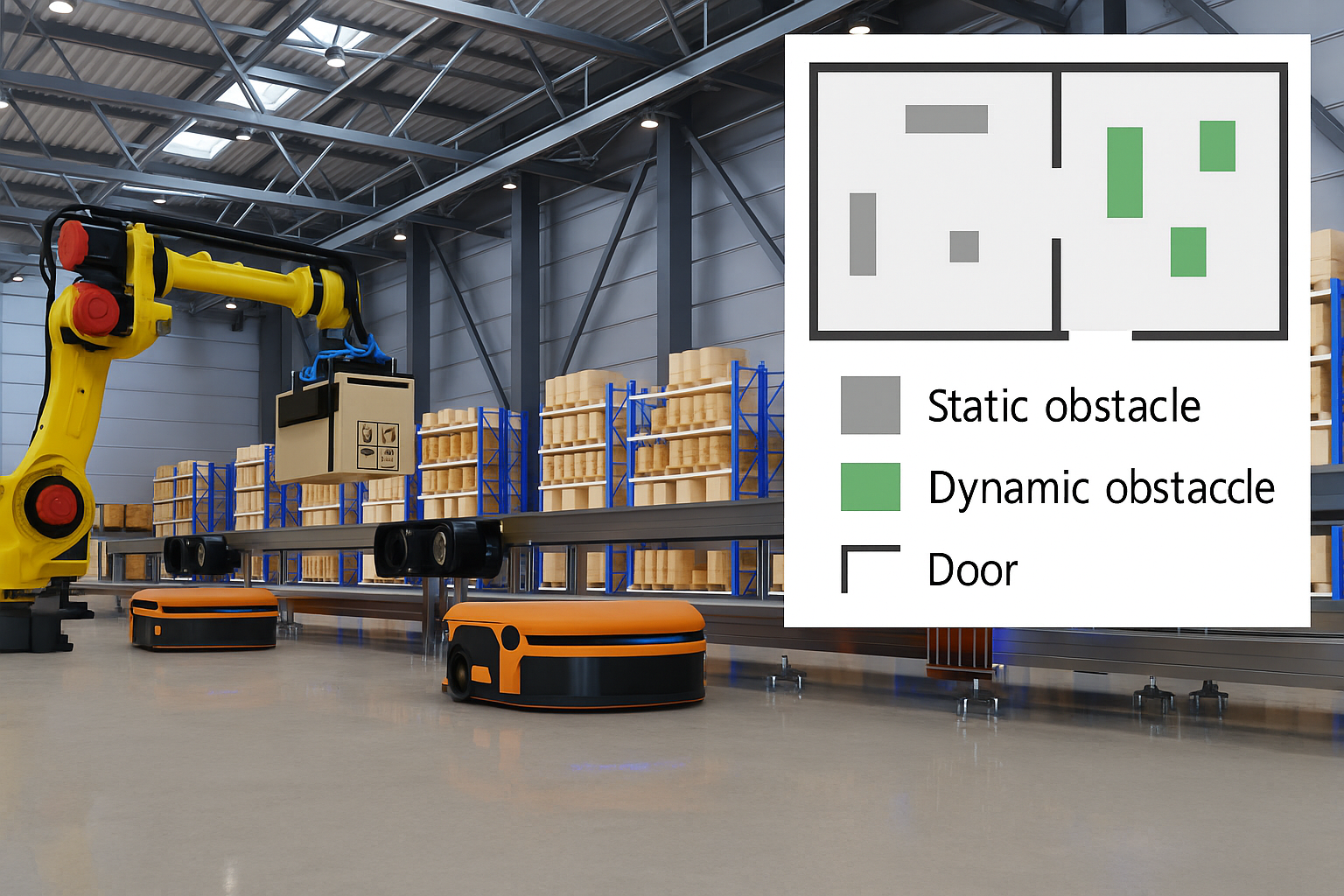}
    \caption{AGV patrolling a warehouse with two partitioned spaces connected by an automated door. The top-right inset map illustrates the environment consisting of static obstacles (e.g., shelving units), and dynamic obstacles (e.g., other AGVs, humans) that may emerge in proximity.}
    \label{fig:case_study}
\end{figure}

\subsection{Safety requirements specification}\label{subsec:SRspec}
In this step, we have to define success and failure criteria through specifying safety requirements. Safety requirements are derived from the hazard of hitting dynamic and static obstacles and ensure a minimum safe distance to them. Furthermore, according to IEC-61058~\cite{IEC61508} SIL 2 specification, a certain level of confidence on this distance is required. Additionally, we have identified a critical element from the Hazard Analysis for when the AGV has to pass through a door:

\begin{itemize}
    \item \textbf{N-SR1}: The AGV shall keep a minimum distance of \(D_{\text{static}}\) to all static objects around it, in all its ODD conditions, at all times.
    \item \textbf{N-SR2}: The AGV shall maintain a minimum separation distance of \(D_{\text{dynamic}}\), from all dynamic objects, including humans, in all its ODM conditions. This value is calculated from the minimum distance required for the AGV to come to a full stop in the worst case scenario where both the vehicle and dynamic object are moving towards each other with their maximum speed.
    \item \textbf{N-SR3}: When passing through narrow passages such as doorways, the AGV shall reduce its velocity.
\end{itemize}

\subsection{Situation Coverage Grid Construction}
We define a structured situation coverage space to systematically explore diverse test scenarios for evaluating the behavior of an autonomous AGV. As shown in Table~\ref{Table:Sit_Grid}, key environmental factors influencing a mission include: whether the AGV must pass through a door, the presence of an obstacle in its path, and the presence of humans or other AGVs in the environment. These four binary factors define the axes of the situation space, resulting in a total of \( 2^4 = 16 \) discrete situations. This exhaustive set is enumerated in Fig.~\ref{fig:situation_coverage_grid_and_DTMC}a, which also presents the expected behavior of the system for each situation (\texttt{s1},\texttt{s2}\dots \texttt{s16}), along with the estimated probabilities of collision with humans and other obstacles. By analyzing these probabilities across different situations, we aim to assess the safety and reliability of the AGV’s navigation and decision-making capabilities in dynamic environments.

\begin{figure*}
    \centering
    \includegraphics[width=1\linewidth]{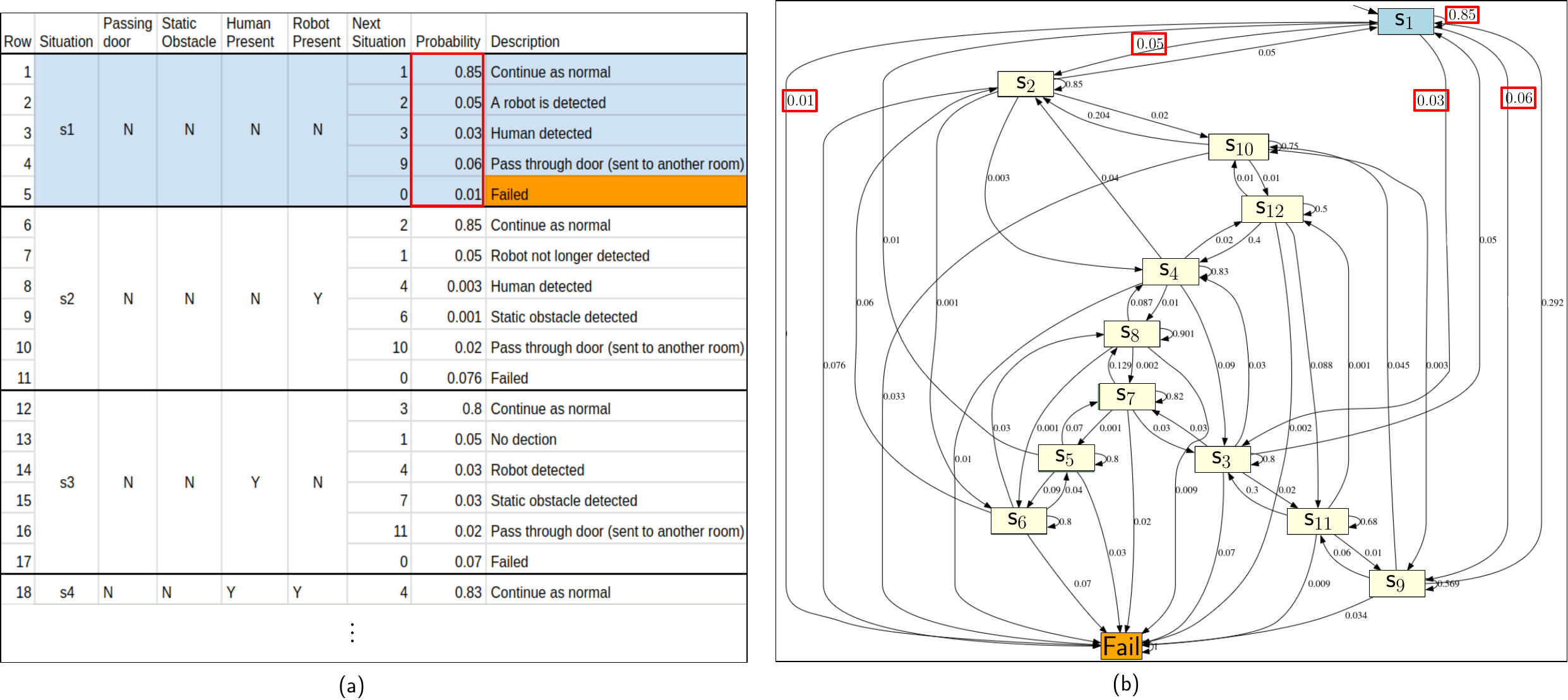}
    \caption{(a)~Augmented Situation coverage grid and (b)~generated DTMC for the AGV scenario. Situation s$_1$ is depicted in blue in both representations, with red rectangles showing its probabilistic transitions. The failure state is depicted in orange.}
    \label{fig:situation_coverage_grid_and_DTMC}
\end{figure*}

\begin{table}[h]
    \centering
    \caption{Situation Coverage Grid}\label{Table:Sit_Grid}
    \begin{tabular}{|p{2.6cm}|p{2.5cm}|p{1cm}|}
    \hline
    \textbf{Axis} & \textbf{Value 1 (description)} & \textbf{Value 2} \\
    \hline
    Passing through door & Yes (crossing door) & No \\
    \hline
    Obstacle on path & Yes (static obstacle detected) & No \\
    \hline
    Human presence & Yes (human detected) & No  \\
    \hline
    AGV presence & Yes (AGV detected) & No \\
    \hline
\end{tabular}
\end{table}

\subsection{Augmenting the Situation coverage Grid}
The situation coverage grid is augmented with quantitative information collected from system tests conducted under different environmental configurations. Changes between situations are observed during testing and modeled as probabilistic distribution functions that capture the likelihood of transitions from one situation to another. For instance, from situation \texttt{s1} (where the AGV encounters no obstacles, humans, or other robots), there is a 0.05 probability that a robot is detected, transitioning to \texttt{s2}, and a 0.03 probability that a human appears, transitioning to \texttt{s3} (see Fig.~\ref{fig:situation_coverage_grid_and_DTMC}a). Similarly, from situation \texttt{s3} (where a human is present), there is a 0.03 chance that another AGV is also detected, resulting in a transition to \texttt{s4}. These transitions enable the generation of a Discrete-Time Markov Chain (DTMC), which models how the system dynamically evolves across situations under uncertain operating conditions.

It has to be noted that, there is a transition to a failure state . This allows to reason about the probability of reaching an unsafe state.
In this example, all failures are grouped together. Different identified failures and easily added into the augmented grid by adding them as new “rows” (i.e., probabilistic transitions into different failures).

Situations that are not observed during the testing stage are excluded from the final coverage grid. While this approach ensures that only empirically encountered scenarios are analyzed, it inherits a key limitation of traditional testing methods: the potential omission of rare but safety-critical situations. Such situations may be theoretically possible within the system’s ODM but are not encountered during the test runs due to their low likelihood. In our case study, we assume that last four situations( \texttt{s13}---\texttt{s16}), specifically those encoded as \texttt{YYNN}, \texttt{YYNY}, \texttt{YYYN}, and \texttt{YYYY} (following the binary encoding convention of presence/absence of environmental factors as defined in Fig.~\ref{fig:situation_coverage_grid_and_DTMC}a)---were not observed during testing. As a result, the total number of covered situations reduces from \( 2^4 = 16 \) to \( 16 - 4 = 12 \) (see Fig.~\ref{fig:situation_coverage_grid_and_DTMC}b) , potentially impacting the completeness of the safety verification.

\subsection{Probabilistic Model Generation}
The system's behavior is represented using a probabilistic state-transition model, where each state (e.g., $S_1$ to $S_{12}$) corresponds to a unique operational situation, and the directed edges indicate transitions between these states with associated probabilities. This model is constructed based on empirical data collected during situation-based testing. For example, starting from state $S_1$, the system has an 85\% probability of remaining in $S_1$, with smaller probabilities of transitioning to $S_2$ (0.05), $S_9$ (0.06), or $S_3$ (0.03), depending on environmental changes and sensor inputs (see Fig.~\ref{fig:situation_coverage_grid_and_DTMC}b). Transitions to a \texttt{Fail} state are explicitly included, enabling quantitative analysis of safety and system reliability. This structure allows for reasoning about the likelihood of reaching unsafe states and supports risk assessment through formal analysis. The model is modular and extensible, allowing new failure modes or behavioral patterns to be integrated as additional probabilistic transitions.

\begin{figure}
    \centering
    \includegraphics[width=.65\linewidth]{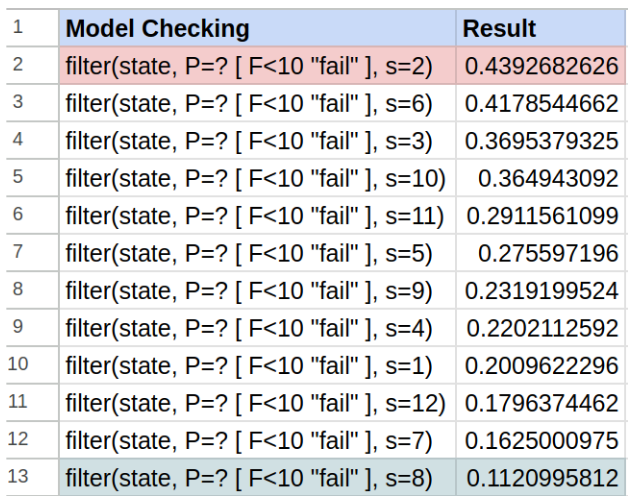}
    \caption{Verification results from each identified situation, ordered from more critical (red) to safer (green) situation.}
    \label{fig:verif_results}
\end{figure}

\subsection{System verification and critical situation analysis}
\label{sec:E System verification}
The quantitative verification of the generated DTMC model under safety properties, previously identified in stage C (subsection \ref{subsec:SRspec}), allows for the identification of the most critical situations (see Fig. \ref{fig:verif_results}). This supports the explainability, decision-making and adaptation process among stakeholders to meet predefined levels of risk. For example, if the maximum probability of failure violates the defined threshold, the system can be modified to increase the probability of success in the most critical situations, or avoid entering such situation all together, if possible.

\section{Conclusion and Future work}

We introduced a structured approach for safety verification of autonomous systems, combining situation coverage with probabilistic modelling and formal verification. Our method captures environmental variability and transition dynamics, enabling the synthesis of models that support quantitative analysis of formally specified safety requirements.

This methodology bridges high-level operational modelling and risk assessment, offering a transparent and explainable tool for safety assurance. Future work will focus on refining environment representation, automating transition probability estimation, and integrating the approach with adaptive control systems. We also aim to evaluate scalability in multi-agent and large-scale real-world domains.



\bibliographystyle{ieeetr}  
\bibliography{references}     

\begin{thebibliography}{10}

\bibitem{PatkarMehendale2025}
V.~Patkar and N.~Mehendale, ``Autonomous ground vehicles: technological advancements, implementation challenges, and future directions,'' {\em International Journal of Intelligent Robotics and Applications}, 2025.
\newblock Received 24 July 2024; accepted 26 December 2024.

\bibitem{Rob2015}
R.~Alexander, H.~R. Hawkins, and A.~J. Rae, ``Situation coverage--a coverage criterion for testing autonomous robots,'' tech. rep., Department of Computer Science, University of York, 2015.

\bibitem{kurakin2018adversarial}
A.~Kurakin, I.~J. Goodfellow, and S.~Bengio, ``Adversarial examples in the physical world,'' in {\em Artificial intelligence safety and security}, pp.~99--112, Chapman and Hall/CRC, 2018.

\bibitem{pei2017deepxplore}
K.~Pei, Y.~Cao, J.~Yang, and S.~Jana, ``Deepxplore: Automated whitebox testing of deep learning systems,'' in {\em proceedings of the 26th Symposium on Operating Systems Principles}, pp.~1--18, 2017.

\bibitem{tian2018deeptest}
Y.~Tian, K.~Pei, S.~Jana, and B.~Ray, ``Deeptest: Automated testing of deep-neural-network-driven autonomous cars,'' in {\em Proceedings of the 40th international conference on software engineering}, pp.~303--314, 2018.

\bibitem{ulbrich2015defining}
S.~Ulbrich, T.~Menzel, A.~Reschka, F.~Schuldt, and M.~Maurer, ``Defining and substantiating the terms scene, situation, and scenario for automated driving,'' in {\em 2015 IEEE 18th international conference on intelligent transportation systems}, pp.~982--988, IEEE, 2015.

\bibitem{abdessalem2018testing}
R.~B. Abdessalem, S.~Nejati, L.~C. Briand, and T.~Stifter, ``Testing vision-based control systems using learnable evolutionary algorithms,'' in {\em Proceedings of the 40th International Conference on Software Engineering}, pp.~1016--1026, 2018.

\bibitem{iqbal2015environment}
M.~Z. Iqbal, A.~Arcuri, and L.~Briand, ``Environment modeling and simulation for automated testing of soft real-time embedded software,'' {\em Software \& Systems Modeling}, vol.~14, pp.~483--524, 2015.

\bibitem{Hawkins2019}
H.~Hawkins and R.~Alexander, ``Situation coverage testing for a simulated autonomous car--an initial case study,'' {\em arXiv preprint arXiv:1911.06501}, 2019.

\bibitem{Tahir2022}
Z.~Tahir and R.~Alexander, ``Intersection focused situation coverage-based verification and validation framework for autonomous vehicles implemented in carla,'' in {\em Modelling and Simulation for Autonomous Systems: 8th International Conference, MESAS 2021, Virtual Event, October 13--14, 2021, Revised Selected Papers}, pp.~191--212, Springer, 2022.

\bibitem{nawshin2023}
N.~M. Proma and R.~Alexander, ``Systematic situation coverage versus random situation coverage for safety testing in an autonomous car simulation,'' in {\em Proceedings of the 12th Latin-American Symposium on Dependable and Secure Computing}, LADC '23, (New York, NY, USA), p.~208–213, Association for Computing Machinery, 2023.

\bibitem{nawshin2024}
N.~Proma, V.~Hodge, and R.~Alexander, ``Situation coverage based safety analysis of an autonomous aerial drone in a mine environment,'' in {\em The Yorkshire Innovation in Science and Engineering Conference (YISEC) 2024}, June 2024.
\newblock This is an author-produced version of the published paper. Uploaded in accordance with the University{\textquoteright}s Research Publications and Open Access policy.

\bibitem{proma2025scaloft}
N.~M. Proma, V.~J. Hodge, and R.~Alexander, ``{SCALOFT: An Initial Approach for Situation Coverage-Based Safety Analysis of an Autonomous Aerial Drone in a Mine Environment},'' in {\em Accepted for, 44th International Conference on Computer Safety, Reliability and Security (safecomp 2025)}, 2025.
\newblock https://arxiv.org/abs/2505.20969.

\bibitem{hawkins2022guidance}
R.~Hawkins, M.~Osborne, M.~Parsons, M.~Nicholson, J.~McDermid, and I.~Habli, ``Guidance on the safety assurance of autonomous systems in complex environments (sace),'' {\em arXiv preprint arXiv:2208.00853}, 2022.

\bibitem{molloy2024hazard}
J.~Molloy, S.~Shahbeigi, and J.~A. McDermid, ``Hazard and safety analysis of machine-learning-based perception capabilities in autonomous vehicles,'' {\em Computer}, vol.~57, no.~11, pp.~60--70, 2024.

\bibitem{ciesinski2004probabilistic}
F.~Ciesinski and M.~Gr{\"o}{\ss}er, ``On probabilistic computation tree logic,'' {\em Validation of stochastic systems: a guide to current research}, pp.~147--188, 2004.

\bibitem{kwiatkowska2022probabilistic}
M.~Kwiatkowska, G.~Norman, and D.~Parker, ``Probabilistic model checking and autonomy,'' {\em Annual review of control, robotics, and autonomous systems}, vol.~5, no.~1, pp.~385--410, 2022.

\bibitem{kwiatkowska2011prism}
M.~Kwiatkowska, G.~Norman, and D.~Parker, ``Prism 4.0: Verification of probabilistic real-time systems,'' in {\em Computer Aided Verification: 23rd International Conferenc (CAV). Proceedings 23}, pp.~585--591, Springer, 2011.

\bibitem{deGelder2024Coverage}
E.~de~Gelder, M.~Buermann, and O.~Op~den Camp, ``Coverage metrics for a scenario database for the scenario-based assessment of automated driving systems,'' vol.~abs/2409.01139, 2024.

\bibitem{iso34504}
``Road vehicles — test scenarios for automated driving systems — scenario categorization,'' 2024.
\newblock First edition, February 2024.

\bibitem{github2025}
Anonymous, ``{Probabilistic Safety Verification}: A situation coverage grid approach.'' \url{https://anonymous.4open.science/r/SafetySituationGrid-14E7}, 2025.
\newblock Anonymous repository.

\bibitem{storm}
C.~Hensel, S.~Junges, J.-P. Katoen, T.~Quatmann, and M.~Volk, ``The probabilistic model checker storm,'' vol.~24, p.~589–610, Aug. 2022.

\bibitem{IEC61508}
{International Electrotechnical Commission}, ``{IEC 61508:2010 – Functional safety of electrical/electronic/programmable electronic safety-related systems}.'' \url{https://webstore.iec.ch/en/publication/5515}, 2010.
\newblock International Standard, Edition 2.0, Parts 1–7.

\end{thebibliography}
\end{document}